\documentclass{article}
\usepackage{spconf,amsmath,amssymb,graphicx}
\usepackage{multirow}
\usepackage{algorithmic}
\usepackage[vlined,ruled]{algorithm2e} 
\SetKwProg{Fn}{Function}{}{}

\title{Can't Fool Me: Adversarially Robust Transformer for Video Understanding}
\name{Divya Choudhary$^*$, Palash Goyal$^*$, Saurabh Sahu$^*$\thanks{$^*$Equal contribution. Names ordered alphabetically}}
\address{Samsung Research America}
\begin{document}
%
\maketitle
\begin{abstract}
Deep neural networks have been shown to perform poorly on adversarial examples. To address this, several techniques have been proposed to increase robustness of a model for image classification tasks. However, in video understanding tasks, developing adversarially robust models is still unexplored. In this paper, we aim to bridge this gap. We first show that simple extensions of image based adversarially robust models slightly improve the worst-case performance. Further, we propose a temporal attention regularization scheme in Transformer to improve the robustness of attention modules to adversarial examples. We illustrate using a large-scale video data set YouTube-8M that the final model (A-ART) achieves close to non-adversarial performance on its adversarial example set. We achieve 91\% GAP on adversarial examples, whereas baseline Transformer and simple adversarial extensions achieve 72.9\% and 82\% respectively, showing significant improvement in robustness over the state-of-the-art.
\end{abstract}

\section{Introduction}
\label{sec:intro}

Deep neural networks have achieved state-of-the-art in several machine learning tasks, e.g., image classification~\cite{kaiming2016image}~\cite{dosovitskiy2020image}, video and audio understanding~\cite{wenjie2018fast},~\cite{esteban2019image},~\cite{tran2019video}, natural language understanding~\cite{jacob2018bert}, graph learning and reinforcement learning~\cite{barret2017reinforcement},~\cite{barret2018reinforcement}. However, it has been shown that due to high dimensionality, even simplest of models are vulnerable to adversarial examples~\cite{carlini2017adv} with imperceptible changes to input examples~\cite{szegedy2013intriguing, 43405, moosavi2016deepfool}.~\cite{wei2019sparse} showed it for videos where the trained model fails to detect the correct class of a perturbed video. For a threat model, they generated the adversarial perturbations in an iterative way by maximising the cross-entropy loss between the model's output for a perturbed video and its ground-truth label. They additionally minimize the $L_{21}$ norm of the perturbations so that the perturbed video is semantically close to the original video. This ensured that visually similar videos had different model outputs.

Adversarial training has been proposed in several key tasks such as image classification to make the model robust to such adversarial changes. For example, in ~\cite{43405, miyato2018virtual}, the authors propose a way of computing the adversarial counterparts of images while training and adding an extra loss regularization that forces the model to correctly classify them or make their predictions close to that of the original image.

However, for video understanding, the research of adversarially robust model needs further exploration. Moving from image to video classification adds several challenges to the task. First, the temporal dimension increases the overall size of the input, in turn increasing the model capacity required to make accurate predictions making it more susceptible to adversarial attacks. Secondly, the number of possible tags increase due to the variations in sequence. For e.g., a leaf falling from tree can be tagged as nature but the reverse could indicate science-fiction elements in the video.

We study the effect of adversarial training on video classification task using a popular deep learning model, Transformer~\cite{vaswani2017attention}, henceforth referred as Non-Adversarially Robust Transformer (Non-ART). For adversarial robustness training, we focus on two major approaches: learning using output space (ART using $\mathcal L_{out, Adv}$) and learning using attention-map space (A-ART using $\mathcal L_{(out,att), Adv}$). We first perform a simple extension (ART) of traditional adversarial loss used in images~\cite{miyato2018virtual} to videos and show that it improves the adversarial robustness to a certain extent. We study the effect of ART in the attention space and show it does not produce an adversarially robust attention space. Based on this, we propose a temporal attention regularization approach (A-ART) and show that it has a large impact on the adversarial robustness of the trained Transformer model. 

We show extensive experiments on a large-scale video data set YouTube-8M on original test set (average performance) as well as adversarial test set (worst-case performance). Further, to showcase generalization, we perform similar experiments on Finance and Arts \& Entertainment verticals within YouTube-8M. We show that we achieve close to average performance on adversarial test set on YouTube-8M using our approach. On original and adversarial test sets, our model (A-ART) achieves 92\% GAP and 91\% GAP respectively. A-ART has a gain of ~18\% in GAP on adversarial test set when compared with the baseline Non-ART and a gain of ~9.1\% when compared with ART. 

\section{Adversarial Transformer}
\label{sec:formu}
In this section, we show how the loss function of a Transformer-based video classification model can be changed for more robust learning. We plan to address the vulnerability of deep learning based video classification models to adversarial examples. We denote the training set with $L$ data points as $\{X_i, y_i\}, i = 1,..,L $, where $X_i \in \mathbb{R}^{T\times D}$ represents the frame-wise feature representation of video $i$ and $y_i \in \mathbb{R}^{K}$ represents the ground-truth labels. $K$ is the number of classes. We represent the video classification model's output vector of probabilities for the point $X_i$ as $\theta (X_i)$.
$l(\theta(X_i), y_i)$ is the loss for the data point $X_i$ which we consider to be cross-entropy loss in our multi-class scenario. 
\begin{equation}
\mathcal L_{CE} = \frac{1}{L} \sum_{i=1}^L l(y_i, \theta(X_i)) \label{eqn:ce}
\end{equation}
However, such models would be susceptible to adversarial examples hurting the generalizability of the model. 
\subsection{Adversarial regularization}
To address the above limitation, we add a regularization term minimizing the loss for adversarial counterparts in the output space of the training samples as proposed in ~\cite{miyato2018virtual}. 
\begin{eqnarray}
\mathcal L_{out, Adv} &=& \mathcal L_{CE} + \alpha * \frac{1}{L} \sum_{i=1}^L l(y_i, \theta(X_i+R_i)) \label{eqn:2}\\
R_i &=& \arg\max_{R_i:\|R_i\|_2\le\epsilon}l(y_i, \theta(X_i+R_i))
\label{eqn:3}
\end{eqnarray}
The loss function is approximated to behave linearly around the input $X_i$ to get the perturbation term which can be easily calculated using backpropagation. 
\begin{equation}
R_i \approx \epsilon \frac{G_i}{\|G_i\|_2}, where\    G_i = \nabla_{X_i}l(y_i, \theta(X_i)) \label{eqn:per}
\end{equation}
Note, this method is closely related to fast gradient sign method (FGSM)~\cite{43405} where $\infty$-norm is considered in equation~\ref{eqn:3}.
We compute the gradient of the loss with respect to the features from video and audio modalities to get the corresponding adversarial perturbations.
Note that, we train the model to be invariant to adversarial
samples within the $\epsilon$ ball (see sec \ref{sec:hyper}). Hence, optimizing this loss function has two hyper-parameters to tune, $\alpha$ and $\epsilon$. We refer to the model trained with $\mathcal L_{out,Adv}$ as Adevrsarially Robust Transformer (ART).

\subsection{Attention-map regularization}
\begin{figure}[t]
\centering
\includegraphics[width=1.0\columnwidth]{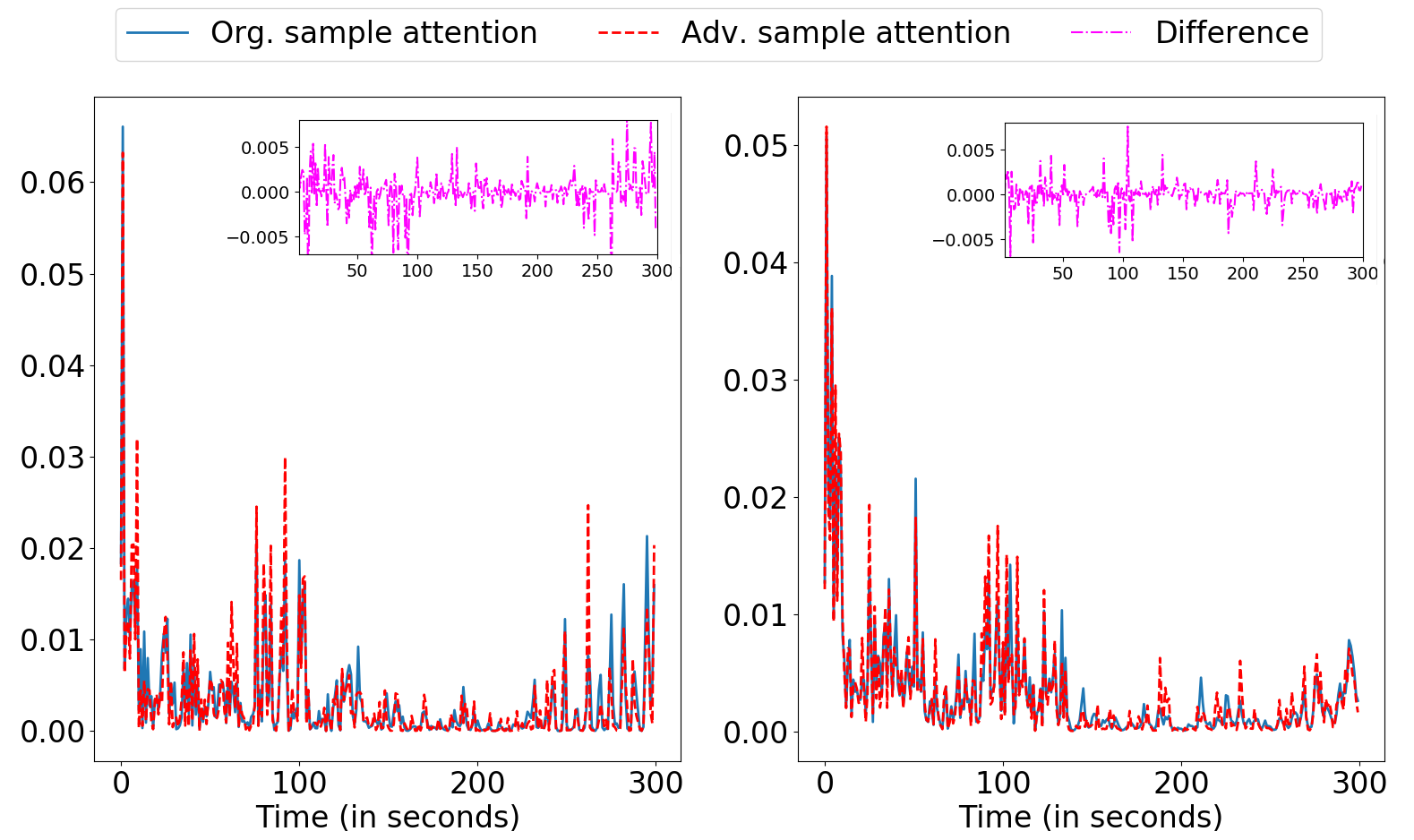}
\caption{Attention map generated by Non-ART (left) and ART (right) models.}
\label{fig:base_att}
\vspace{-4mm}
\end{figure}
Figure ~\ref{fig:base_att} illustrates attention map generated for a sample video and its adversarial counterpart by Non-ART (model trained with $\mathcal L_{CE}$)  and ART model. We observe that even though the attention maps are close, for many frames the differences are still significant. But intuitively, the attention-map being generated in the self-attention blocks should also be invariant to adversarial examples. Towards that goal, we add a regularization term in our loss function to enforce this condition. For a given input, we average over the attention maps $A_m$ generated by each head to get the attention map $A \in \mathbb{R}^{T\times T}$ for that input. The corresponding attention map generated using the adversarial example is denoted by $A^{adv}$. We minimize the Frobenius norm of the difference between the attention maps and average over the mini-batch in which case the loss function becomes
\begin{equation}
\mathcal L_{(out,att), Adv} = \mathcal L_{out, Adv} + \beta_{Fr} * \frac{1}{L} \sum_{i=1}^L \|A_i - A^{adv}_i\|_{Fr} \nonumber \\
\end{equation}
We refer to the model trained with $\mathcal L_{(out,att), Adv}$ as Attention-Adversarially Robust Transformer (A-ART)

\section{Experiments}
\label{sec:exp}


\begin{table*}[!htbp]
  \centering
  \scriptsize
  \begin{tabular}{l||llll||llll}
    \hline
       & \multicolumn{4}{c}{Test} & \multicolumn{4}{c}{Adversarial Test} \\
    \hline
    Metrics   &  Non-ART, $\mathcal L_{CE}$   & ART, $\mathcal L_{out,Adv}$ & A-ART, $\mathcal L_{(out,att), Adv}$& $Gain_{Test}$  &  Non-ART, $\mathcal L_{CE}$   & ART, $\mathcal L_{out,Adv}$ & A-ART, $\mathcal L_{(out,att), Adv}$ & $Gain_{Adv}$\\
    \hline
    GAP       &    91.97$\pm$0.02 &  92.38$\pm$0.02 & 92.00$\pm$0.02 &\textbf{+0.03} & 72.89$\pm$0.04 & 81.97$\pm$0.02 & 91.04$\pm$0.02 & \textbf{+18.15}             \\ 
PERR     &   89.73$\pm$0.04 & 89.94$\pm$0.04& 89.64$\pm$0.04 & \textbf{-0.09}  & 71.89$\pm$0.03 & 80.35$\pm$0.03 & 88.58$\pm$0.01 & \textbf{+16.69}         \\
Hit@1    &  94.76$\pm$0.05  & 94.87$\pm$0.04& 94.80$\pm$0.03 & \textbf{+0.04}   & 81.74$\pm$0.02 & 88.22$\pm$0.02 & 93.99$\pm$0.02 & \textbf{+12.25}      \\\hline
  \end{tabular}
\caption{Overall performance (in percentages) of Non-ART, ART and A-ART models for video categorization task on our YouTube-8M test sets. Each column represents the results for a training paradigm defined by the architecture and the loss function used. We present the mean and standard deviation obtained for five non-overlapping partitions of the entire test set.}
\label{tab:res}
\end{table*}

\begin{table*}[!htbp]
  \centering
  \scriptsize
  \begin{tabular}{l||llll||llll}
    \hline
       & \multicolumn{4}{c}{Test} & \multicolumn{4}{c}{Adversarial Test} \\
    \hline
    Metrics   &  Non-ART, $\mathcal L_{CE}$   & ART, $\mathcal L_{out,Adv}$ & A-ART, $\mathcal L_{(out,att), Adv}$& $Gain_{Test}$  &  Non-ART, $\mathcal L_{CE}$   & ART, $\mathcal L_{out,Adv}$ & A-ART, $\mathcal L_{(out,att), Adv}$ & $Gain_{Adv}$\\
    \hline
    GAP &  93.93 &  94.12 & 94.14 &\textbf{+0.21} & 74.88 & 84.38 & 88.54 & \textbf{+13.66}  \\ 
    PERR  &   93.47 & 93.72 & 93.57 & \textbf{+0.10}  & 78.40 & 85.60 & 88.25  & \textbf{+9.85}  \\
    Hit@1    &  97.21  & 97.32 & 97.33 & \textbf{+0.01}   & 87.38 & 92.16 & 93.95 & \textbf{+6.57} \\
    \hline
    \end{tabular}
\caption{Overall performance (in percentages) of baselines and different variations of our proposed model for video categorization task on \textit{Arts \& Entertainment}, largest vertical of YT8M.}
\label{tab:Arts}
\end{table*}

\begin{table*}[!htbp]
  \centering
  \scriptsize
  \begin{tabular}{l||llll||llll}
    \hline
       & \multicolumn{4}{c}{Test} & \multicolumn{4}{c}{Adversarial Test} \\
    \hline
    Metrics   &  Non-ART, $\mathcal L_{CE}$   & ART, $\mathcal L_{out,Adv}$ & A-ART, $\mathcal L_{(out,att), Adv}$& $Gain_{Test}$  &  Non-ART, $\mathcal L_{CE}$   & ART, $\mathcal L_{out,Adv}$ & A-ART, $\mathcal L_{(out,att), Adv}$ & $Gain_{Adv}$\\
    \hline
    GAP &  79.13 &  79.07 & 79.86 & \textbf{+0.73} & 54.50 & 65.08 & 71.48 & \textbf{+16.98}  \\ 
    PERR  &   80.59 & 83.85 & 82.25 & \textbf{+1.66}  & 64.77 & 70.89 & 75.50  & \textbf{+10.73}  \\
    Hit@1    &  87.70  & 92.31 & 90.77 & \textbf{+3.07}   &  69.25  & 74.77 & 85.60 & \textbf{+16.35} \\
    \hline
  \end{tabular}
\caption{Overall performance (in percentages) of baselines and different variations of our proposed model for video categorization task on \textit{Finance}, smallest vertical of YT8M.}
\label{tab:Finance}
\vspace{-0.4cm}
\end{table*}

\subsection{Experimental-setup}
We use YouTube-8M dataset for our experiments which consists of frame-wise video and audio features for approximately 5 million videos extracted using Inception v3 and VGGish respectively followed by PCA ~\cite{abu2016youtube}. We use the hierarchical label space with 317 classes, a further modification of 431 categories (see \cite{sahu2020cross}). We use binary cross-entropy loss to train our models. We evaluate our models using the three metrics mentioned in ~\cite{lee20182nd}: (i) Global Average Precision (GAP), (ii) Precision at Equal Recall Rate (PERR), and (iii) Hit@1. 
Our training set consists of approximately 4 million videos. We use 64000 videos from the official development set for validation and use the rest as test set. Our baseline Transformer model consists of a single layer of multi-head attention with 8 attention heads for each of audio and video modalities. For training we used Adam optimizer, with an initial learning rate of 0.0002 and batch size of 64. We compute validation set GAP every 10000 iterations and perform early-stopping with patience of 5.
We also use it for learning-rate scheduler that decreases the learning rate by a factor of 0.1 with patience of 3.

\subsection{Results}
We present results for video categorization using a baseline Transformer encoder with and without the proposed regularization terms. Based on validation set results, the value of $\epsilon$ and $\alpha$ for adversarial training was set to be $0.5$ and $1$ respectively. $\beta_{Fr}$ was set to $0.001$.
We first analyze the effect of attention map regularization on ART's performance on the test set consisting of original samples as well as adversarial samples generated using the perturbations computed as in equation~\ref{eqn:per}. Then we perform hyper-parameter analysis of the models trained adversarially.  Finally, we highlight the robustness of A-ART to adversarial perturbations computed using the DeepFool method~\cite{moosavi2016deepfool}.

\subsubsection{Performance on original and adversarial samples}
From Table \ref{tab:res}, we see that adversarial regularisation improves the performance of Non-ART model when classifying the original samples in test set. This shows that adversarial regularization can improve the generalizability of a model. A-ART and ART seem to perform similarly on the original test set. Next, we investigate the robustness of models to adversarial samples. Given a test sample and a trained model, we generate the corresponding adversarial perturbation using the FGSM based method (equation~\ref{eqn:per}). We observe that A-ART consistently and significantly outperforms ART highlighting the importance of attention-map resgularization to improve the  adversarial robustness of Transformer based models.

Hence, while A-ART performs similarly as ART on the original test samples, it significantly improves the adversarial robustness of the model.

\subsubsection{Attention Map Regularization}
\renewcommand{\thefigure}{2}
\begin{figure}[hbtp]
\centering
    \includegraphics[width=1.0\columnwidth]{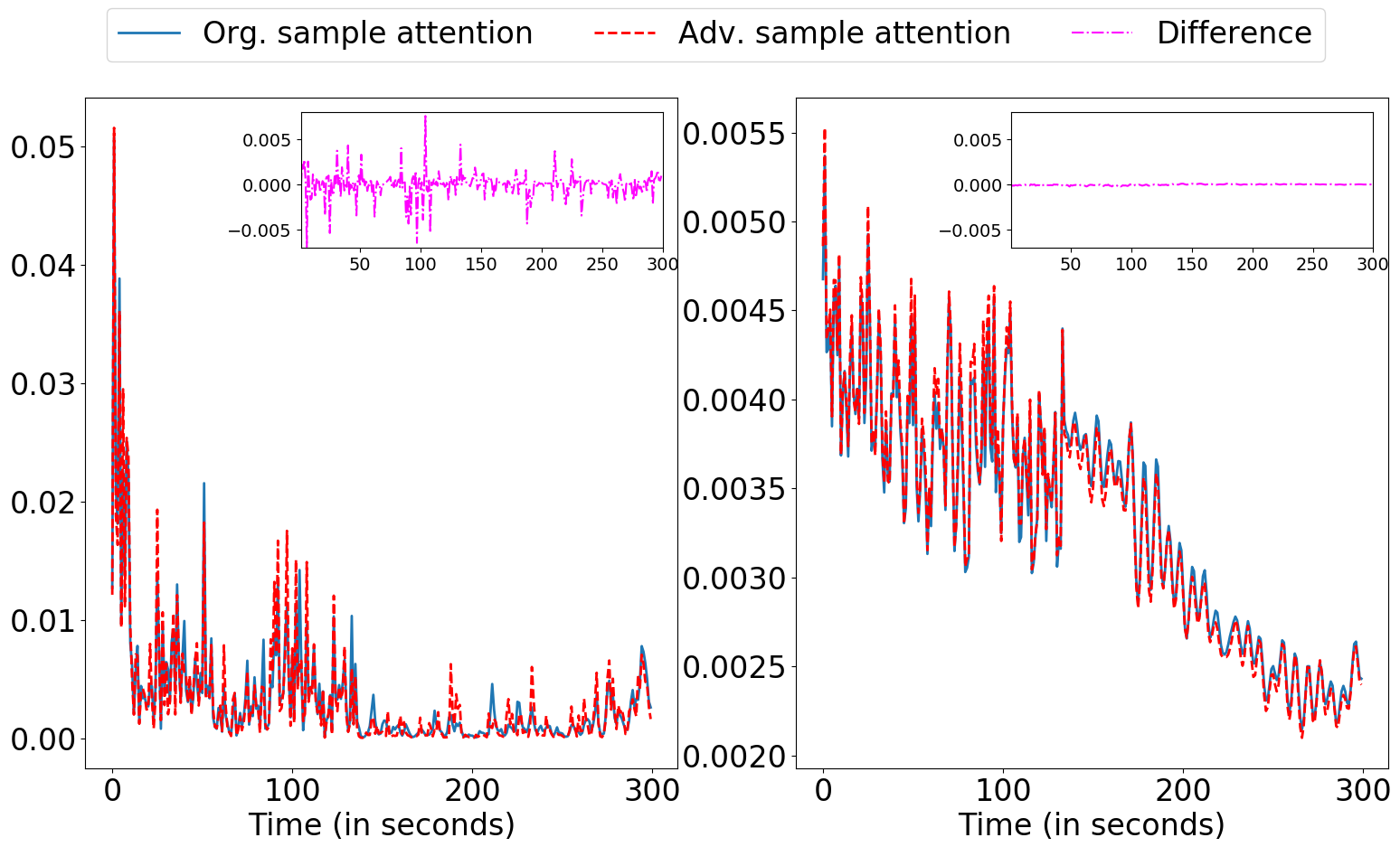}
    \caption{Attention profile generated by ART (left) and A-ART (right) for a test sample and its adversary.}
    \label{fig:ATvsATFr2}
    \vspace{-0.3cm}
\end{figure}

In Figure~\ref{fig:ATvsATFr2}, we compare the attention profiles being generated for a sample in test set and its adversarial counterpart by trained ART and A-ART models. We observe that that attention generated by A-ART is more robust to adversarial perturbations and the attention profiles of the original sample and its adversarial counterpart overlap to a great extent. On the other hand, ART exhibits more variations in the attention profiles as a result of adversarial perturbations to the input. Another thing to notice from the figure is that the maximum attention being given to any frame reduces by an order of magnitude when trained using A-ART model. In other words, the attention map generated by A-ART is smoother enforcing the temporal coherence property which has been shown to help video classification model performance~\cite{huang2018makes, mobahi2009deep}.

For the test set videos, we computed the mean square error (MSE) between the attention-maps obtained for the original video and its adversarial counterpart using ART and A-ART. Then we averaged it over the entire test set. We note that our proposed model, A-ART drastically reduced the average MSE from $4 \times 10^{-5}$ obtained using ART to $3 \times 10^{-8}$. This ensures that A-ART has learnt to not to be fooled by perturbations in the adversarial example and has almost similar attention-map for both the original and perturbed video.

\subsubsection{Hyperparameter Tuning}
\label{sec:hyper}
\renewcommand{\thefigure}{3}
\begin{figure}[hbtp]
\centering
    \includegraphics[width=0.95\columnwidth]{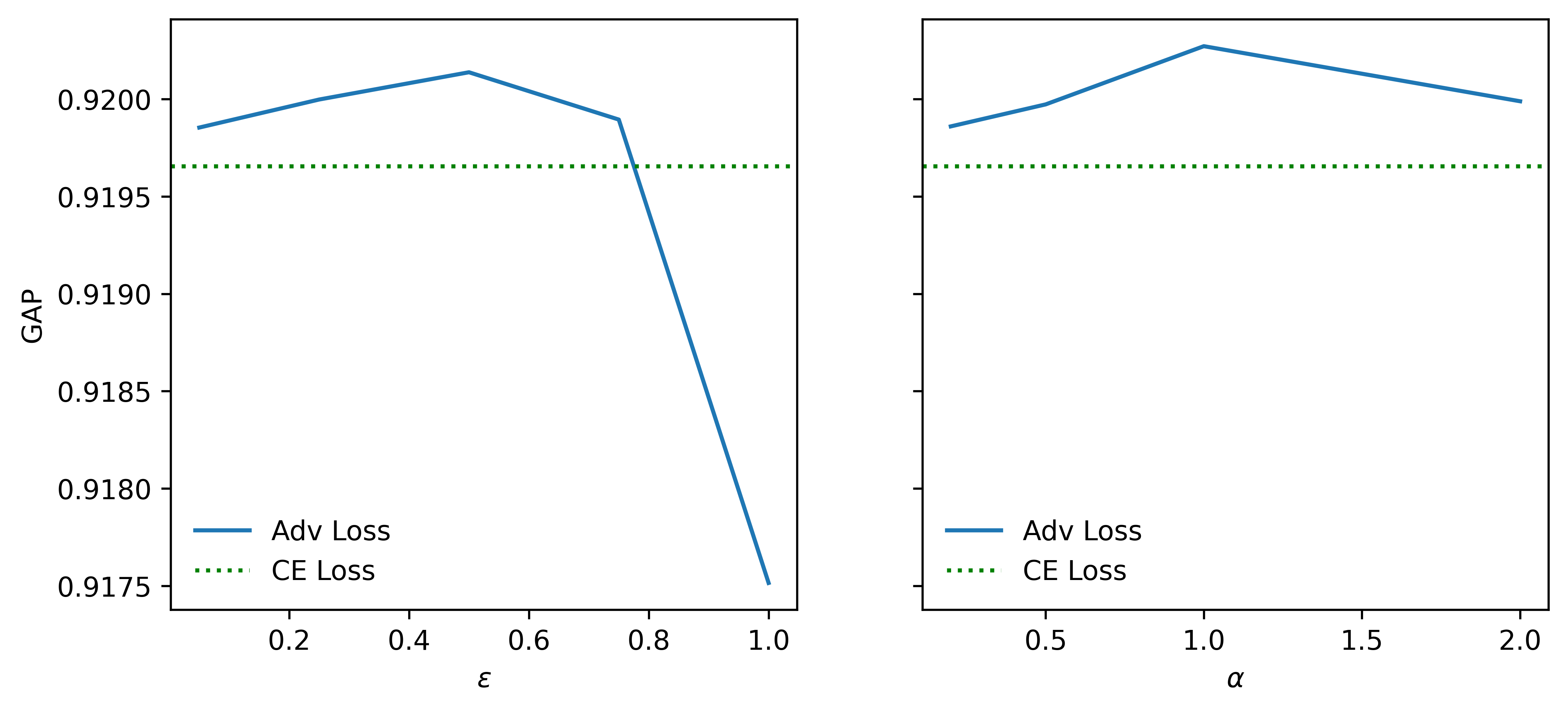}
    \caption{Validation GAP for models with cross-entropy loss and adversarial loss: neighborhood radius $\epsilon$ was varied with $\alpha = 1$(left) and $\alpha$ was varied with $\epsilon = 0.5$ (right).}
    \label{fig:AT}
    \vspace{-0.2cm}
\end{figure}

We aim to understand the impact of the two hyper-parameters $\epsilon$ and $\alpha$ on the model performance by perturbing one of them, while keeping the other constant. By altering $\epsilon$ , we aim to understand the impact of smoothing radius around the data-points on the model performance and perturbing $\alpha$ impacts the weight of the adversarial loss on the overall optimization. The plots comparing the validation GAP for different values of hyper-parameters is shown in Figure~\ref{fig:AT}. First, the value of $\alpha$ was kept fixed at 1 and $\epsilon$ was varied. For lower values of $\epsilon$, ART model show an improvement over Non-ART model peaking at $\epsilon=0.5$.  As we increase the value of $\epsilon$, the model’s performance starts deteriorating. This is expected since $\epsilon$ defines the neighborhood around an input feature vector over which the conditional distribution is smoothed. Increasing the radius of this neighborhood forces our model to learn smoother functions that cannot capture the complexity of the conditional distribution function thereby decreasing its performance on the validation set. Similarly, as we increase the adversarial loss weight $\alpha$, the performance increases, peaks at $1$ and starts reducing as the relative weight of classification goes down.

\subsubsection{Adversarial Robustness}
Moosavi et.al \cite{moosavi2016deepfool} proposed a simple and accurate method for computing the robustness of different classifiers to adversarial perturbations. Given a data sample and a trained model, their method computes the minimum perturbation $r_{tot}$ that can be added to the sample so that the model predicts it incorrectly. Then a statistic $\rho_{tot}$ is computed by dividing the norm of $r_{tot}$ by the norm of actual features. To fool an adversarially robust model, it would require a greater amount of perturbation to be added to the features. This means $\rho_{tot}$ should be higher for a more robust model. We compute $\rho_{tot}$ using our trained models and YT8M test set samples and show the results in Fig \ref{fig:robust}. It can be seen clearly that A-ART improves the adversarial robustness of ART.
\renewcommand{\thefigure}{4}
\begin{figure}[hbtp]
\centering
    \includegraphics[width=0.95\columnwidth]{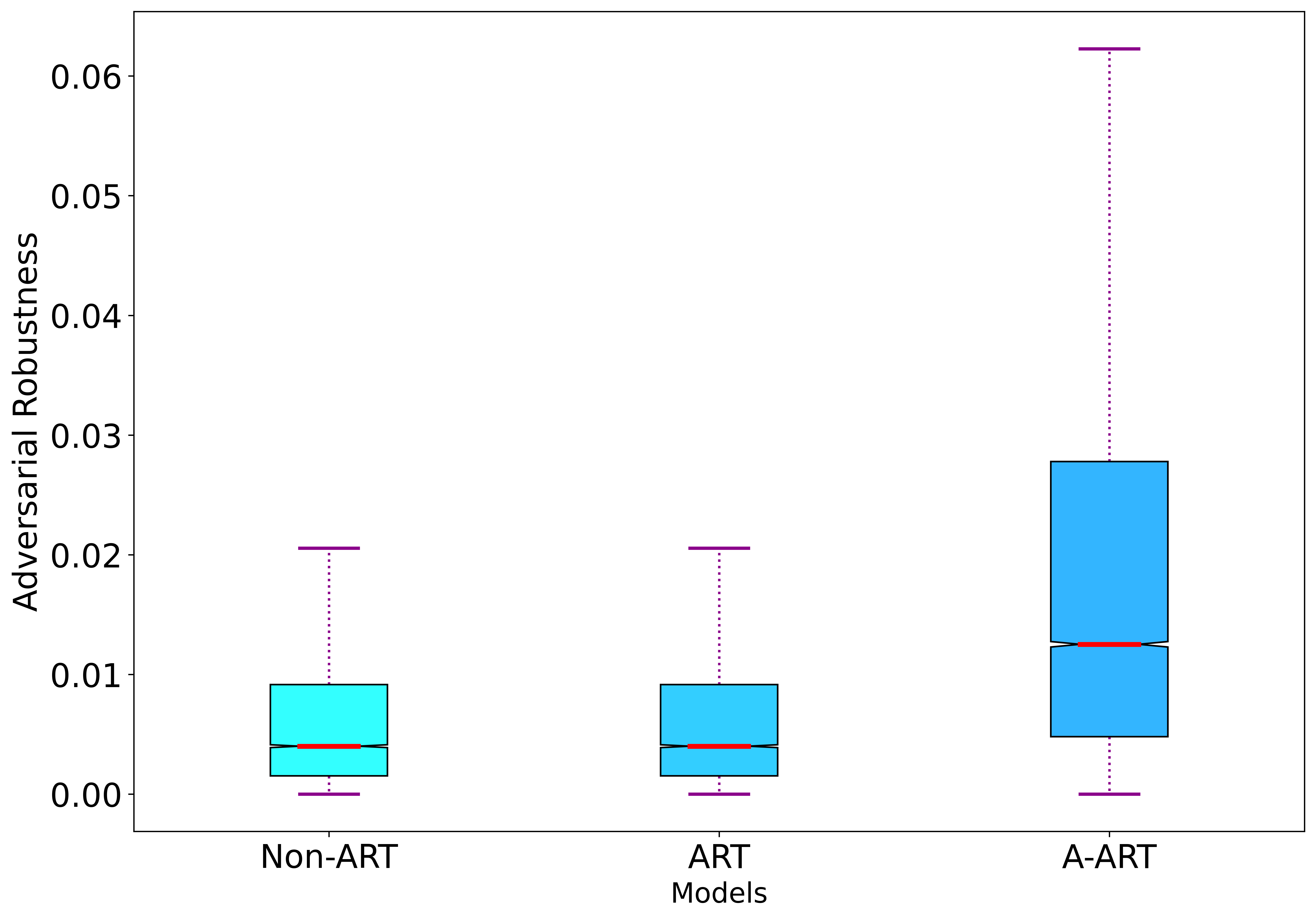}
    \caption{Average Robustness for all models. Our proposed model A-ART is more robust to adversarial examples than both ART and non-ART base model.}
    \label{fig:robust}
    \vspace{-0.4cm}
\end{figure}



\section{Conclusion}
\label{sec:conclu}
This paper presents two approaches to train adversarially robust Transformer model: (i) ART, an extension of image-based adversarially robust model for videos and (ii) A-ART, an approach to further improve robustness of attention space as well as output space. We show that ART and A-ART perform better than Non-ART on test set. Moreover, compared to Non-ART and ART, A-ART shows an exceptional gain of 18\% and 9\% respectively in robustness to adversarial examples generated using FGSM based method. We also show enhanced robustness of A-ART to adversarial perturbations generated using DeepFool. We also observed that the attention-map generated by A-ART is more robust to adversarial perturbations. In future, we plan to investigate the robustness of intermediate embeddings so that they can be used to improve other video understanding tasks. We also plan to extend A-ART to raw video datasets to train and qualitatively evaluate on more realistic adversarial examples.

\vfill\pagebreak

\bibliographystyle{IEEEbib}
\bibliography{bib}

\end{document}